%% file: cvpr.tex
\documentclass[final]{cvpr}

\usepackage{times}
\usepackage{epsfig}
\usepackage{graphicx}
\usepackage{amsmath}
\usepackage{amssymb}
\usepackage{subfigure}

\usepackage[pagebackref=true,breaklinks=true,colorlinks,bookmarks=false]{hyperref}

\def\cvprPaperID{****} 
\def\confYear{CVPR 2021}

\begin{document}

\title{Good Practices and A Strong Baseline for Traffic Anomaly Detection}
\author{Yuxiang Zhao~$^{1,2}$\thanks{The first two authors contributed equally to this work. This work was done when Yuxiang Zhao was a research intern at Baidu.}\quad
Wenhao Wu~$^{2*}$\quad
Yue He~$^{2}$\quad
Yingying Li~$^{2}$\quad
Xiao Tan~$^{2}$\quad
Shifeng Chen~$^{1}$\thanks{Corresponding author.}\\
$^1$ Shenzhen Institute of Advanced Technology, Chinese Academy of Sciences, China\\
$^2$ Department of Computer Vision Technology (VIS), Baidu Inc., China\\
}

\maketitle
\begin{abstract}

The detection of traffic anomalies is a critical component of the intelligent city transportation management system. Previous works have proposed a variety of notable insights and taken a step forward in this field, however, dealing with the complex traffic environment remains a challenge. 
Moreover, the lack of high-quality data and the complexity of the traffic scene, motivate us to study this problem from a hand-crafted perspective. 
In this paper, we propose a straightforward and efficient framework that includes pre-processing, a dynamic track module, and post-processing. 
With video stabilization, background modeling, and vehicle detection, the pro-processing phase aims to generate candidate anomalies. 
The dynamic tracking module seeks and locates the start time of anomalies by utilizing vehicle motion patterns and spatiotemporal status. 
Finally, we use the post-processing to fine-tune the temporal boundary of anomalies. 
Not surprisingly, our proposed framework was ranked $1^{st}$ in the NVIDIA AI CITY 2021 leaderboard for traffic anomaly detection. 
The code is available at: \url{https://github.com/Endeavour10020/AICity2021-Anomaly-Detection }.
\end{abstract}

\section{Introduction}

\begin{figure}[t]
    \centering
    \includegraphics[width=0.98\linewidth]{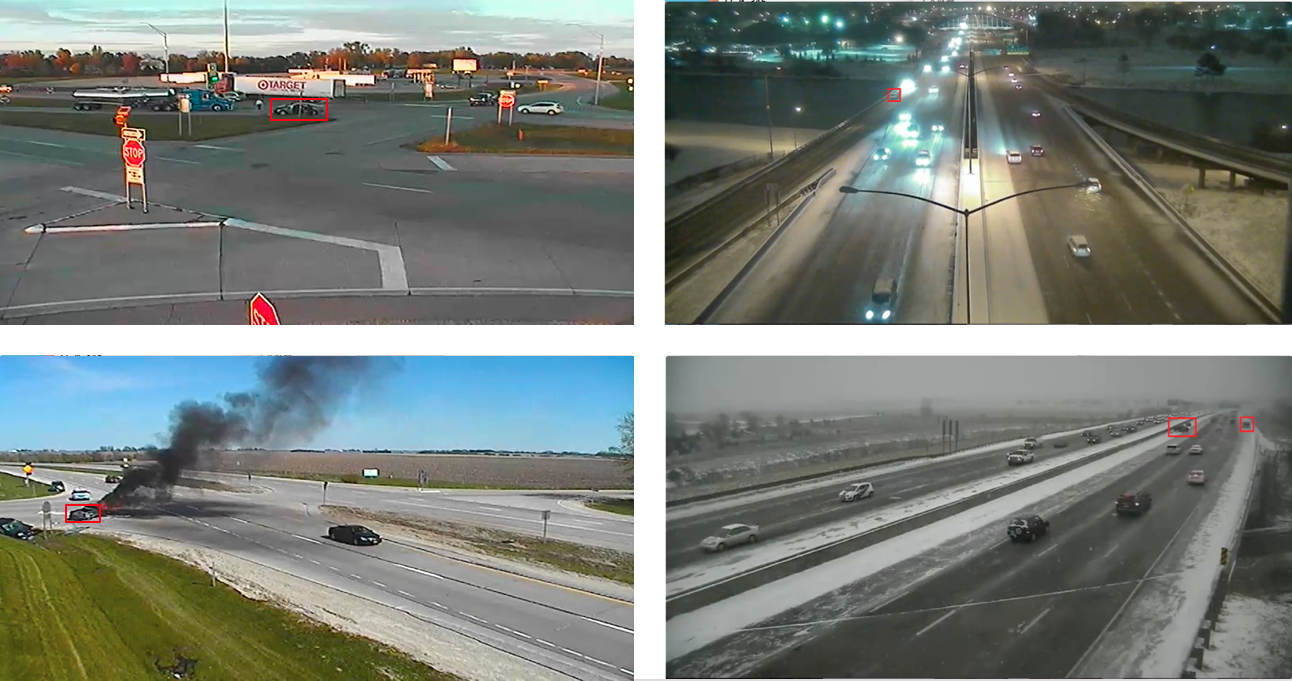}
    \caption{Examples of traffic anomalies. \textcolor[RGB]{246,0,0}{Red rectangles} highlight the abnormal instances.}
    \label{fig:figure1}
\end{figure}

With the ongoing expansion of urban traffic system construction, large-scale traffic management systems and accident warning systems have become an essential component of urban infrastructure development. 
Traffic anomaly detection is a critical task in video understanding, which has attracted widespread attention from both academia and industry.
As shown in Figure \ref{fig:figure1}, due to the complexity of the traffic scene and the diversity of camera views, there remain great challenges in traffic anomaly detection.

Generally, anomalies rarely occur as compared to normal activities. Hence, developing intelligent computer vision methods for automatic video anomaly detection is a pressing need.
The goal of a practical anomaly detection system is to timely signal an activity that deviates normal patterns and identifies the time window of the occurring anomaly. Therefore, anomaly detection can be considered as coarse-level video understanding, which filters out anomalies from normal patterns. 

Recently, deep neural networks have achieved great success in image recognition~\cite{simonyan2014very,he2016deep,chen2017deeplab}, video understanding~\cite{wu2019multi,wu2020dynamic,wu2020MVFNet,wu2021dsanet,wang2021temporal}, object detection~\cite{ren2015faster,redmon2018yolov3,liu2016ssd}, \emph{etc}. 
Deep ConvNets come with excellent modeling capacity and are capable of learning discriminative representations from visual data in large-scale supervised datasets (\emph{e.g.,} ImageNet~\cite{deng2009imagenet}, Kinetics-400~\cite{kay2017kinetics}, Places~\cite{zhou2017places}, MS COCO~\cite{lin2014microsoft}).
However, unlike these above tasks, the application of end-to-end ConvNets to traffic anomaly detection is impeded by the following major obstacle: Training deep ConvNets framework in an end-to-end manner usually requires a large volume of training samples to achieve optimal performance. However, the training set of Track4 only consists of 100 videos, which remain limited in both size and diversity.

\begin{figure*}[t]
    \centering
    \includegraphics[width=0.98\linewidth]{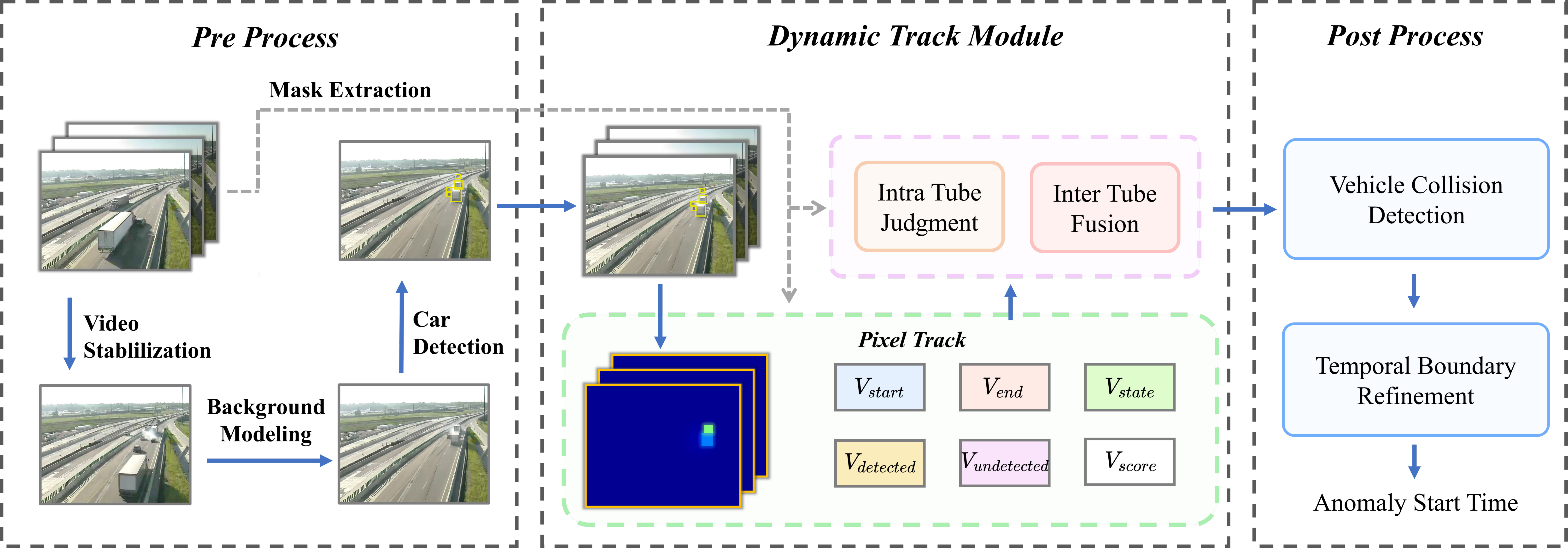}
    \caption{The illustration of hierarchical strategy framework pipeline. This framework includes pre-processing, dynamic tracking module, and post-processing.}
    \label{fig:framework}
\end{figure*}

These challenges motivate us to study traffic anomaly detection from a hand-crafted perspective. 
In this paper, we develop a customized architecture with human prior knowledge, which provides a conceptually simple, and robust framework for traffic anomaly detection.
As shown in Figure \ref{fig:framework}, our framework consists of pre-processing, dynamic tracking module, and post-processing. 

Concretely, the pipeline of pre-processing including vehicle detection, video stabilization, background modeling. 
The pro-processing phase aims to generate the candidate anomalies with video stabilization, background modeling, and vehicle detection.
The dynamic tracking module utilizes the motion patterns and spatio-temporal status of vehicles to seek and locate the start time of the anomalies. 
Finally, we employ the post process to further refine the temporal boundary of anomalies.
The experimental results show that our proposed framework is effective and robust in the Track4 test set of the NVIDIA AI CITY 2021 Challenge, which ranks first place in this competition with a 95.24\% F1-score and 5.3080 root mean square error, and the final score is 93.55\%.


The main contributions are summarized as follows:
\begin{itemize}
    \item We present a hierarchical architecture with human prior knowledge, which provides a conceptually simple framework for traffic anomaly detection.
    \item We investigate a series of good practices for temporal localization and anomaly detection (\emph{e.g.,} video stabilization, vehicle collision detection, temporal boundary refinement, \emph{etc.}).
    \item Results demonstrate that our method outperforms other challengers on the Track 4 test set of the NVIDIA AI CITY 2021 Challenge. 
\end{itemize}

\section{Related Work}

Anomaly detection is the process of detecting rare or unusual patterns that deviate from the normal behavior, which is called ``Outliers" or ``Anomalies".
Many studies have explored this field with a variety of methods for a long time. These approaches can be divided into two parts by the types of models, traditional methods and deep learning based methods. Some works use 
traditional machine learning models like Gaussian mixture models \cite{li2016traffic}, Regression models \cite{cheng2015gaussian}, histogram-based \cite{zhang2016combining}, Dirichlet process mixture models (DPMM) \cite{ngan2015outlier}, Bayesian network-based models \cite{blair2014event} to seek anomalies. With the great development of computer vision leveraged on deep learning, autoencoder based networks and well-designed loss becomes the key point of anomaly forecasting task \cite{medel2016anomaly,li2016anomaly,patil2016global,wang2018abnormal,wang2014detection,chang2018video,tan2016fast,yu2016content,zhang2015abnormal,liu2018accumulated,mo2013adaptive,chen2015detecting,wang2018automatic}. Sultani \emph{et al.}~\cite{sultani2018real} construct a new large-scale dataset, called UCF-Crime, which contains real-world anomaly videos.

Traffic anomaly detection is a more fine-grained anomaly detection, which includes multiple kinds of violations of regulations such as driving in the wrong direction, illegal parking,\emph{etc}.

In the past years of NVIDIA AI CITY Challenges, unsupervised traffic anomaly detection methods have become the mainstream in real traffic accident scenarios and promote the development of intelligent transportation. Wei \emph{et al.} ~\cite{wei2018unsupervised} first remove the moving vehicles using MOG2 while keeping the stopped vehicles as part of the background, then perform multi-scale detection and classification to detect anomalies. Shine \emph{et al.} ~\cite{shine2019comparative} show an unsupervised method to tackle this problem including a background extraction stage, an anomaly detection that identifies the stalled vehicles in the background, and a final anomaly confirmation module. Bai \emph{et al.} ~\cite{bai2019traffic} present a novel spatial-temporal information matrix, which transforms the analysis of a strip trajectory into an analysis of the spatial position. Shine \emph{et al.}~\cite{shine2020fractional} conduct background subtraction using GMM and utilize YOLO detector to select anomaly candidates, then detect anomalies by transfer learning without using training data  ~\cite{doshi2020fast}. Li \emph{et al.}~\cite{li2020multi} present a multi-granularity tracking approach, which combines a box-level branch and a  pixel-level branch to analyze the candidate abnormal vehicles at different granularity levels. They ranked first in NVIDIA AI CITY 2020.

In this paper, we propose a simple and effective framework for traffic anomaly detection in track 4 of the NVIDIA AI CITY 2021 Challenge. We obtain a 0.9524 F1-score with a detection time error of just 5.3080 seconds and our results rank first place on the track 4 test set among all the participant teams.

\begin{figure}[t]
    \centering
    \includegraphics[width=0.98\linewidth]{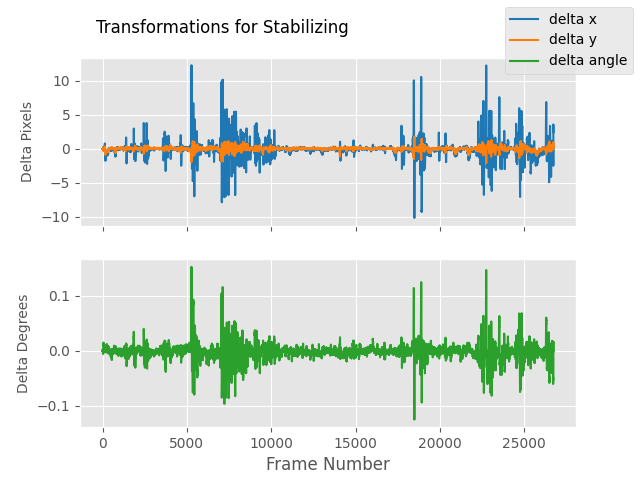}
    \caption{Visualization of video stabilization.}
    \label{fig:DVS}
\end{figure}





\section{Method}
In this section, we first present the pipeline of pre-processing, which includes video stabilization, background modeling, vehicle detection, and mask generation.
Then we illustrate our dynamic track module to locate the start time of the anomaly. 
Finally, we depict our post-processing pipeline which aims to further refine the temporal boundary of anomalies.

\begin{figure}[t]
    \centering
    \subfigure[Forward background modeling.]{
    \includegraphics[width=0.98\linewidth]{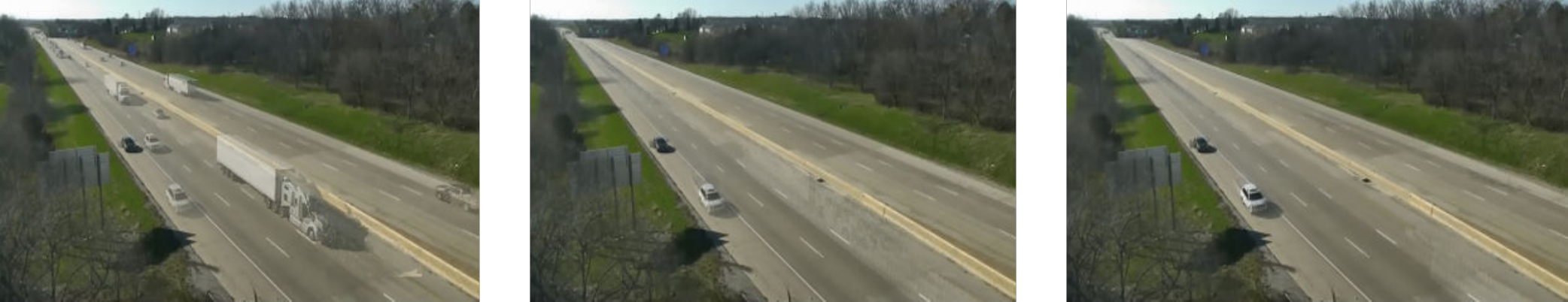}
    }
    \subfigure[Backward background modeling]{
    \includegraphics[width=0.98\linewidth]{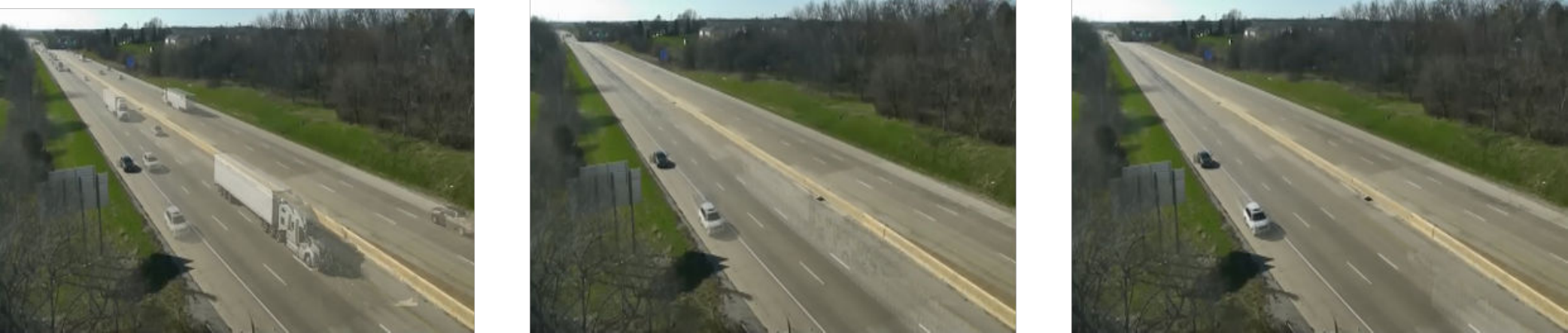}
    }
    \caption{Examples of background modeling.}
    \label{fig:BG}
\end{figure}

\subsection{Pre-Processing}
\paragraph{Video Stabilization.}
In a real scenario, the traffic camera under adverse conditions usually results in non-precise motion and occurrence of shaking, which may compromise the stability of the obtained videos, and directly influence the performance of the modules (\emph{e.g.}, detection, background modeling) in our framework.

To overcome such a problem, we perform digital video stabilization (DVS) to correct camera motion oscillations that occur in the acquisition process, through software techniques, without the use of specific hardware, to enhance visual quality and improve final applications, such as vehicle detection and tracking of vehicles.
The whole pipeline of DVS could be roughly divided into two stages: The first stage is to estimate the camera movements. Then the second stage is to correct and smooth the camera motion.

Considering the moving vehicles may affect the quality of motion detection, we apply a robust video stabilization method to compensate for the undesired movements of the camera.
To be specific, we utilize the combination of feature point matching based on GFTT~\cite{GFTT} and calculate sparse optical flow to generate frame-to-frame transformations. A hybrid filter was used in the video stabilization process to smooth the trajectory of transformations.
Figure~\ref{fig:DVS} shows the variety of $\delta_{x}$, $\delta_{y}$ and $\delta_{angle}$ in transformation matrix. By using these parameters, we can filter the shacking heavily videos that exceed accumulated threshold $\delta_{t}$ and average threshold $\delta_{avg}$ to imply video stabilization.

\paragraph{Background Modeling.}
It is common knowledge that vehicles involved in a traffic accident or anomaly will often come to a complete stop naturally. 
As a result, the traffic anomaly detection task is transformed into static vehicle detection. This allows us to effectively detect anomalies of the stopped vehicle on the lane, even when the videos are of poor quality.

Various works in the field of object tracking have attempted to distinguish the foreground and background. Due to the poor video sources, these methods require a robust and adaptive background representation. 
To dynamically model backgrounds, we use the background modeling approach based on the mixture of Gaussians (MOG) \cite{zivkovic2004improved}.
We perform an ablation analysis on the duration of background modeling to make a trade-off between effectiveness and efficiency. 
We compare the difference between the forward and backward background modeling results in Figure~\ref{fig:BG}, which shows that backward background modeling results will make stopped vehicles clearer, and we use the forward background modeling approach as an auxiliary method to get a more accurate start time of the anomaly.

\begin{figure}[t]
    \centering
    \includegraphics[width=0.98\linewidth]{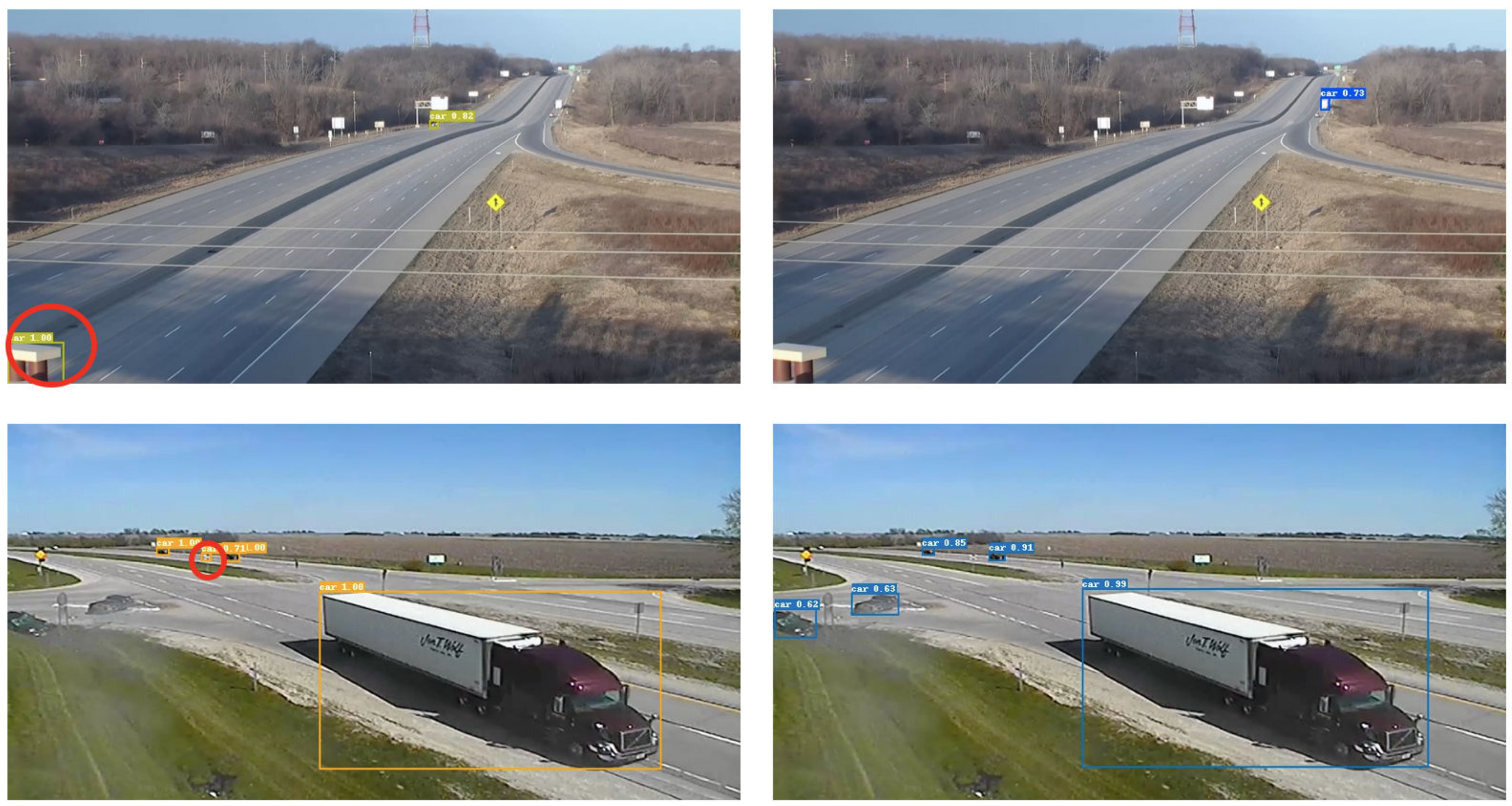}
    \caption{Visualization of vehicle detection. Left is the results of Faster RCNN and right is results of Cascase RCNN. \textcolor[RGB]{246,0,0}{Red circles} highlight the false detection.}
    \label{fig:det}
\end{figure}



\paragraph{Vehicle Detection.}
For object detection, there are two mainstream methods as follows: one-stage detectors and two-stage detectors.
Two-stage detectors are more accurate and require a higher computational time. On the contrary,  one-stage detectors benefited from straightforward architectures, which are faster and simpler but might potentially drag down the performance.
To obtain robust detection results, in this paper, we build a two-stream vehicle detection module that consists of two two-stage detectors: Faster R-CNN \cite{ren2015faster} and Cascade R-CNN~\cite{cai2018cascade}.
Concretely, we employ the Faster R-CNN with SENet-152 \cite{hu2018squeeze}  as our main detector for vehicle detection.
The Cascade R-CNN with CBResnet-200 \cite{liu2020cbnet}, is used as an auxiliary detector to reduce the fail of vehicle detection.
Following the common practice, we apply the Feature Pyramid Network (FPN) \cite{lin2017feature} to build high-level semantic feature maps at all scales.
Figure~\ref{fig:det} shows the results of our detector.

\paragraph{Mask Generation.}
In general, traffic anomalies usually happen on vehicles driving the main road.
Hence, we need to filter the static vehicles on the side roads and parking lots.
A practicable solution is to segment out hypothetical abnormal mask regions automatically.
To do so, image segmentation is an intuitive approach to distinguish the hypothetical abnormal area. However, due to the complexity of the road scene and limited training videos, it is hard to learn a robust segmentation model.

\begin{figure}[t]
    \centering
    \includegraphics[width=0.98\linewidth]{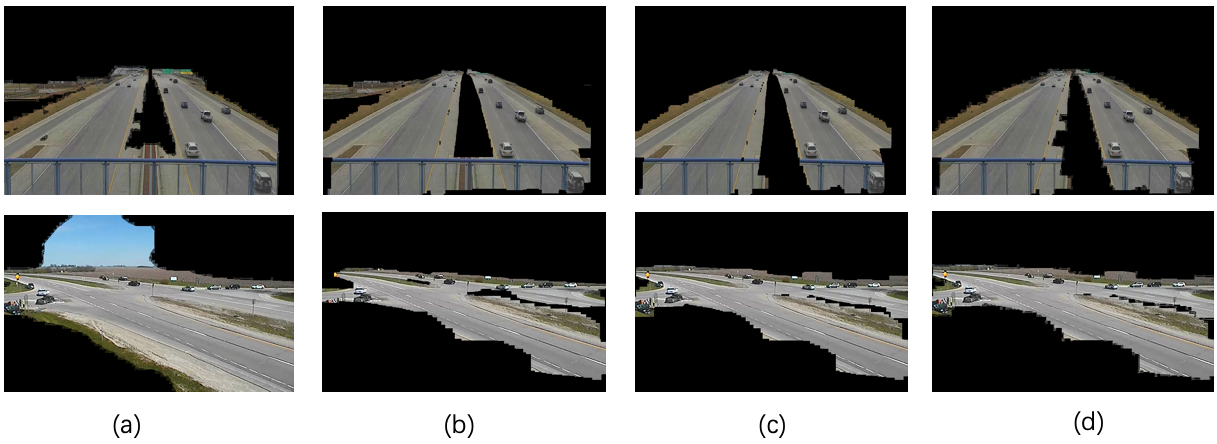}
    \caption{Examples of the abnormal mask. From the left to the right: motion-based mask, original trajectory-based mask, our enhanced trajectory-based mask, and final fused mask. Our trajectory-based mask effectively reduces false auxiliary road recall, and the abnormal area adjacent to the main road is reserved.}
    \label{fig:mask}
\end{figure}

Inspired by the motion-based mask extraction method in \cite{li2020multi}. Then, we suggest an enhanced trajectory-based mask solution to produce finer masks. 
To obtain the vehicle's trajectory, we use the Track1 dataset to train the multi-object tracking algorithm DeepSORT~\cite{deepsort}. We cluster these trajectories into the primary and secondary parts by calculating the angle of moving direction, which is different from the previous process. The bounding box of the detected vehicle will be used to expand the trajectory-based mask if the absolute value of $\arctan \theta$ is in the primary part.
The camera orientation is taken into account in this way, resulting in a more reasonable traffic road mask. 
Finally, we combine the two masks above to get the final result. These two masks can be used in conjunction, and Figure \ref{fig:mask} demonstrates some results of the abnormal mask.

\subsection{Dynamic Tracking Module}

\paragraph{Pixel Tracking.}
Inspired by \cite{li2020multi}, we adopt pixel-level information to filter out suspicious events as much as possible. Six spatial-temporal matrics, including   $V_{undetected}$, $V_{detected}$, $V_{score}$, $V_{state}$, $V_{start}$ and $V_{end}$, are combined to record anomaly car detection results iteratively. As shown in Figure \ref{fig:framework},
the pixel region and time stamp of the detection will be continuously accumulated on the six matrixes when a car appears on the background frame. 
If the peak temporal span exceeded the time threshold, the suspicious results will be thrown out. In addition, we jointly develop two mechanisms to optimize the anomaly detection result as follows:
1) When an abnormal object is observed, we will utilize the IOU algorithm to compare the current object position with the detection result in the next frame.
When the intersection of the two regions is greater than 0.5, we update the start timestamp of an anomaly.
2) We develop a backtracking approach based on the spatial-temporal correlation similarity since certain cars do not stop immediately in an accident. When IoU is less than the threshold, the PSNR and color histogram features are extracted for the non-overlapped bounding boxes to calculate box similarity.

Then, in the following phase, we begin to process the candidate results. First, we filter the true anomaly start time using intra tube judgment. Second, we use inter tube fusion to join tubes from the same vehicle together.


\begin{figure}[t]
    \centering
    \includegraphics[width=0.98\linewidth]{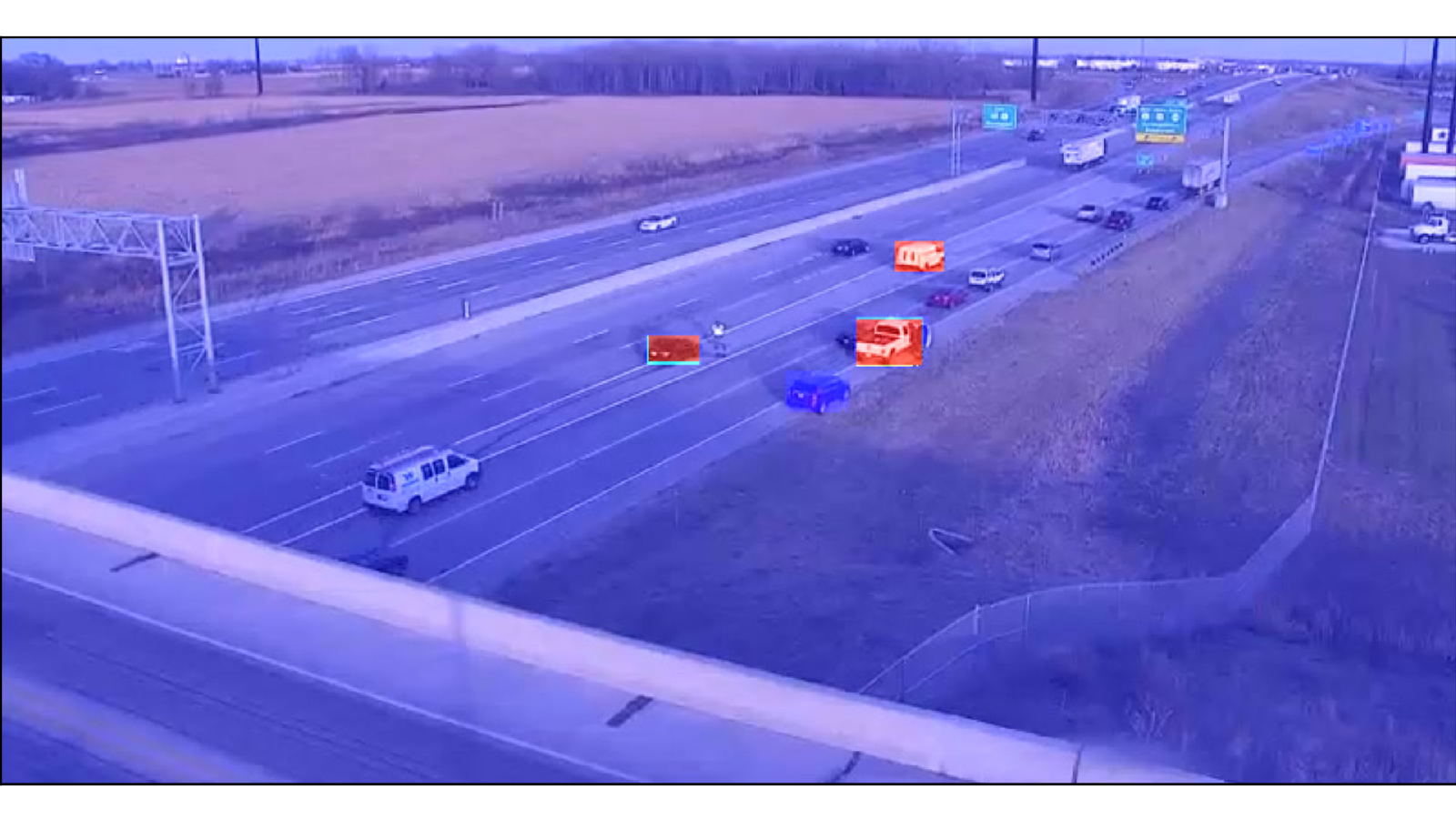}
    \caption{Visualization of Pixel track.}
\end{figure}

\begin{figure*}[t]
    \centering
    \subfigure[Video 33 on test set]{
    \includegraphics[width=0.98\linewidth]{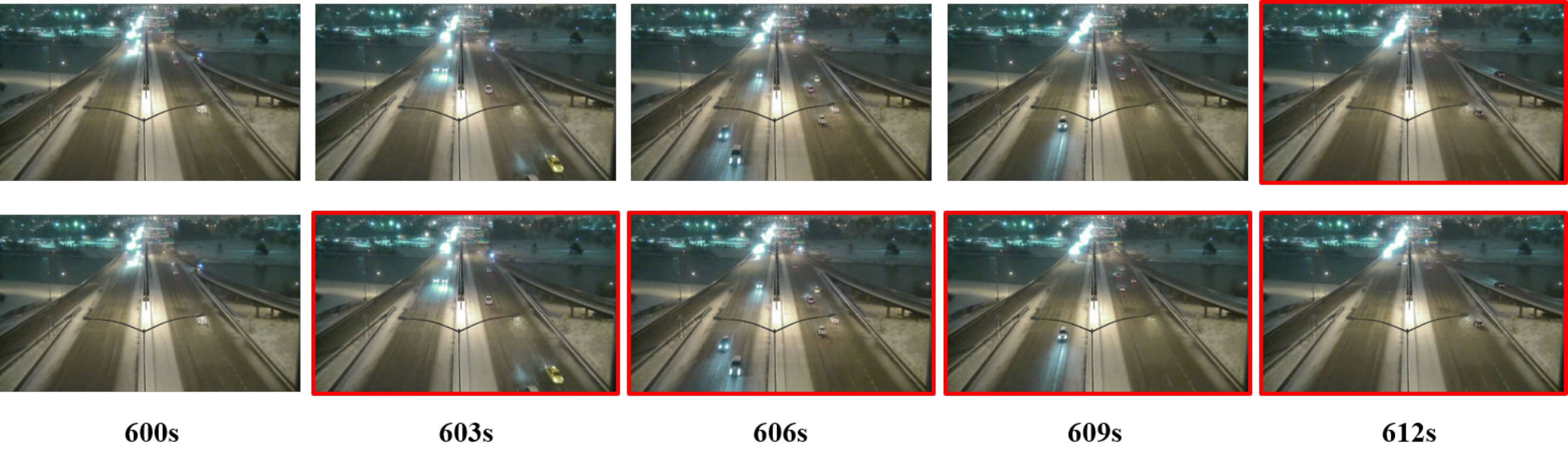}
    }
    \subfigure[Video 73 on test set]{
    \includegraphics[width=0.98\linewidth]{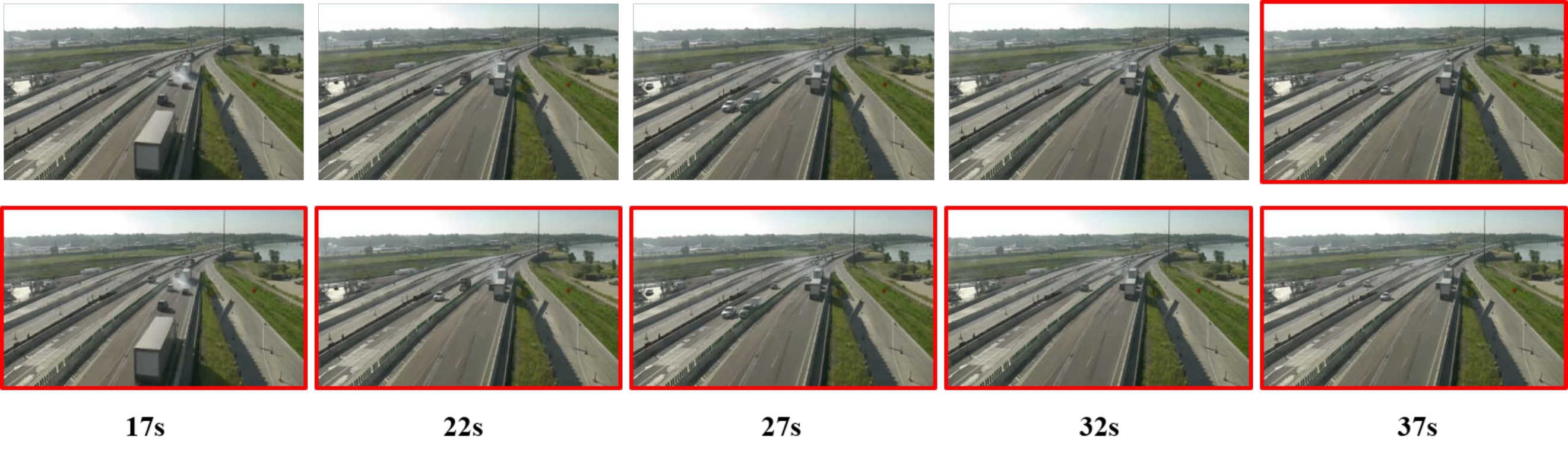}
    }
    \caption{Example results of vehicle collision detection. For video 33, we backtrack the start time from the stopped time (612 seconds) to the crash time (603 seconds). For video 73, we backtrack the start time from the delayed stopped time (37 seconds) to the crash time (17 seconds).}
    \label{fig:crash_det}
\end{figure*}

\paragraph{Intra tube judgment.}
The candidate tube may contain several vehicles, making it difficult for the model to determine the true anomaly start time. So we conduct intra tube judgment to remove the parts that do not belong to the current anomalous vehicle. Here we denote the $i$-th tube as $U^{i}$, and the $j$-th frame of it as $U^{i}_{j}$. ${Reg}^{i}$ are regions of tube $U^{i}$ in chronological order. The start time of tube $U^{i}$ is denoted by $U_{Start}^{i}$, the end time is denoted by $U_{End}^{i}$. The 
regions in ${Reg}^{i}$ across the timeline are composited by the set $S({Reg}^{i})$. We first calculate the similarity between every 
element in $S({Reg}^{i})$ and the mean over the interval $\overline{{U^{i}}}$, and store it in the set $S(Sim_{j}^{i})$. We use SSIM as the similarity measure which ranges from $(0, 1)$.

Then we compute mean-subtracted value $S'(Sim_{j}^{i})$ for every element in $S(Sim_{j}^{i})$. We judge whether the region in $j$-th frame belongs to the current anomalous car or not in this manner: 
$$ \left\{\begin{array}{ll}
S'(Sim_{j}^{i})>{Thre}_{Sim},\ False \\
LowerBound<S'(Sim_{j}^{i})\leq {Thre}_{Sim},\ True
\end{array}\right.  $$
That is to say, the $S'(Sim_{j}^{i})$ which exceeds the threshold $Thre_{Sim}$ demonstrates that $Reg^{i}_{j}$ is not the bounding box of the current anomalous car on the global time. This comparison is performed in chronological order until $Sim_{j}^{i}$ is lower than ${Thre}_{Sim}$ and $j$-th timestamp is recorded as a potential start of an anomaly. If a successive satisfied duration from $j$-th timestamp exceeds $\gamma (U_{End}^{i}-U_{Start}^{i})$, then $j$-th timestamp is assigned as the starting time of the true anomaly. $\gamma$ is a hyperparameter that controls the tolerance. This process should be continued until all the candidate tubes are judged.


\paragraph{Inter tube fusion.}
One vehicle can appear in two tubes. In order to fuse candidate tubes that represent the same vehicle, we perform the following steps to fuse these tubes. Generally, one vehicle that appeared in two spatial locations in a picture is highly similar in appearance and bounding box scale. Based on this hypothesis, we compute the similarity between the mean of region sets of two tubes,  $Sim(\overline{U^i},\overline{U^k})$. If the similarity results exceed a defined threshold, the initial start time and end time are $min(U_{Start}^i, U_{Start}^j)$ and $max(U_{End}^i, U_{End}^j)$ respectively.

\subsection{Post Processing}
\paragraph{Vehicle Collision Detection.}
Stalled vehicles and crashes are the most common anomalies in the training set. 
The anomaly start time for a stalled vehicle is when the vehicle comes to a complete stop. The methods described above are capable of accurately predicting the stalled vehicle's abnormal start time. 
The start time for a single-vehicle crash or a multiple-vehicle crash is the moment the first crash happens rather than the time the vehicle comes to a complete stop. As a result, the above method's prediction would be somewhat delayed. 
To that end, we devise a method for detecting vehicle collisions.
First, we obtain the estimated time when the vehicle comes to a full stop. Then, along the temporal axis, we trace this vehicle backwards. 

By analyzing the videos of crashes, we find that there will be obvious changes in the foreground at the moment of a collision on the surrounding areas of the wrecked car. Meanwhile, the background image of the surrounding areas will be different between the time of the crash and after the vehicle has stopped completely. Based on the above, we can judge the accurate collision time of vehicles.
The results is shown in Figure \ref{fig:crash_det}.

\paragraph{Temporal Boundary Refinement.}
Forward background modeling will cause the anomaly to be displayed as afterimages after a delay, causing a temporal prediction delay. To obtain a more accurate temporal localization, we use reverse background modeling and appearance similarity to refine the anomaly's start and end times.

\section{Experiments}
NVIDIA AI CITY 2021 Challenge presents a dataset of 100 videos for training and 150 videos for testing. 
The dataset includes events that occurred naturally or abnormally in a variety of severe weather conditions, such as raining, snowing, or fogging. Each video is about 15 minutes long and has a resolution of 800$\times$410 pixels. 
In this section, we first introduce the experimental setup, then we provide the implementation details. Finally, we describe the evaluation metrics and present the results of our method.

\subsection{Implementation Details}
\paragraph{Vehicle Detection.}
To train the Faster R-CNN and Cascade R-CNN, we use SGD with momentum 0.9 and weight decay 1e-4. 
With an initial learning rate of 0.01 and a minibatch of 8, the Faster R-CNN model is trained for 50K iterations. 
The learning rate is reduced by a factor of 10 at iteration 30K and 40K. Besides, the Cascade R-CNN model is trained with an initial learning rate of 0.005 and reduced by a factor of 10 at iteration 40K and 60K.
To obtain more data for training, following the official instructions, we use data across different challenge tracks (\emph{i.e.,} Track1 and Track3) to pre-train our detectors, then we finetune the model with the training videos of Track4.
The shorter side of the input images is resized to 800 pixels while keeping the aspect ratio.
We use 5 scale anchors of \{$16^2, 32^2, 64^2, 128^2, 256^2$\} and ground-truth boxes are associated with anchors, which have been assigned to pyramid levels.
The NMS is used to filter the final vehicle detection result, which is made up of the union of two models with a threshold of 0.8.
Moreover, we train our models with the PaddlePaddle deep learning framework~\footnote{https://github.com/paddlepaddle/paddle}.

\paragraph{Video Stablization.}
For distinguishing a shacking video, the accumulated threshold $\delta_{t}$ is set to 17200 for video stabilization, and the average threshold $\delta_{avg}$ is set to 0.645.

\paragraph{Road Mask.}
Following the protocol in \cite{li2020multi}, we use the same parameters in motion-based mask. In our enhanced trajectory-based mask, the minimum trajectory length $n$ is set to 5, and the minimum distance $d$ of the trajectory is set to 50. 
Further, the threshold of the angle of moving direction to distinguish primary and secondary parts is set to 0.8.

\paragraph{Pixel Tracking.}

The thresholds for the normal-suspicious and suspicious/abnormal-normal state transitions are all set to three consecutive frames. 
The time thresholds for suspicious and coarse anomaly candidates have been set to 20s and 30s, respectively. 
The relaxed constraint satisfaction ratio $T_{ratio}^r$ is 0.6 and the shortest traceback time $T_{time}$ is 30s. The IoU thresholds $T_{IoU}$ and $T_{IoU}^r$ are 0.3 and 0.5, respectively; the PSNR thresholds $T_{PSNR}$ and $T_{PSNR}^r$  are 18 and 20, respectively; and the color histogram thresholds $T_{Color}$ and $T_{Color}^r$ are 0.88 and 0.9, respectively.

\paragraph{Intra Tube Judgment \& Inter tube fusion.}

In Intra tube, the min and max threshold of ssim similarity $Thresh_{similarity}$ is set to 0.25 and 0.6 respectively. The duration time ratio $\gamma$ is fixed to 0.3. In inter tube, the PSNR threshold between two tubes is set to 18.

\paragraph{Vehicle Collision Detection.}
The threshold of the surrounding area is set to 50, the threshold of foreground changes is set to 1000 and the similarity threshold of the background image is 0.9.


\subsection{Evaluation Metrics}
Evaluation for Track 4 will be based on model anomaly detection performance, measured by the F1-score, and detection time error, measured by RMSE. 
Specifically, the track 4 score will be computed as:
  \begin{equation}
S4 = F1 \times (1-\mathrm{NRMSE}),
 \end{equation}
where F1 is the F1-score and NRMSE is the normalized root mean square error (RMSE). The S4 score ranges between 0 and 1, and higher scores are better.

The F1-score is computed as Eq.~\ref{f1}.
For the purpose of computing the F1-score, a true-positive (TP) detection will be considered as the predicted anomaly within 10 seconds of the true anomaly (i.e.,  seconds before or after) that has the highest confidence score. Each predicted anomaly will only be a TP for one true anomaly. A false-positive (FP) is a predicted anomaly that is not a TP for some anomaly. Finally, a false-negative (FN) is a true anomaly that was not predicted.

\begin{equation}
F_{1}=2 \cdot \frac{\text { Precision } \cdot \text { Recall }}{\text { Precision }+\text { Recall }} \\
=\frac{\mathrm{TP}}{\mathrm{TP}+\frac{1}{2}(\mathrm{FP}+\mathrm{FN})}.
\label{f1}
\end{equation}

We compute the detection time error as the RMSE of the ground truth anomaly time and predicted anomaly time for all TP predictions. In order to eliminate jitter during submissions, normalization will be done using min-max normalization with a minimum value of 0 and a maximum value of 300, which represents a reasonable range of RMSE values for the task.
NRMSE is the normalized root mean square error (RMSE) as follow:
\begin{equation}
 \mathrm{NRMSE} = \frac{\min( \sqrt{\frac{1}{\mathrm{TP}}\sum_{i=1}^{\mathrm{TP}}(t_i^{p}-t_i^{gt})^2}, 300)}{300}.
 \end{equation}

\subsection{Experiments results}
We evaluate our model on NVIDIA AI CITY 2021 Challenge test dataset. 
The F1-score is 0.9524, and the RMSE is 5.3080, as shown in Table \ref{ours}.
The results demonstrate the effectiveness and robustness of our method.
The leaderboard for all teams is shown in Table~\ref{rank}, and we are in \textbf{first} place with a score of 0.9355.

\begin{table}[h]
\centering
\caption{Our results on Track4 test set.}
\vspace{5pt}
\begin{tabular}{c|c|c}
\hline
F1     & RMSE   & S4 Score \\ \hline
0.9524 & 5.3080 & 0.9355 \\ 
\hline
\end{tabular}
\label{ours}
\end{table}

\input{tab/rank}

\section{Conclusion}
In this paper, we propose a simple and effective framework for detecting traffic anomalies (\emph{e.g.,} stalled car, car crash).
In Track 4 of the NVIDIA AI CITY 2021 Challenge, we obtain a 0.9524 F1-score with a detection time error of just 5.3080 seconds. 
Our results rank first place on the Track 4 test set among all the participant teams, which demonstrates that the superiority of our method.

\section*{Acknowledgment}
 This work is supported by Key-Area Research and Development Program of Guangdong Province (2019B010155003), Shenzhen Science and Technology Innovation Commission (JCYJ20200109114835623), National Natural Science Foundation of China (U1713203), and the  Scientific  Instrument  Developing project of the Chinese Academy of Sciences (YJKYYQ20190028).

{\small
\bibliographystyle{ieee_fullname}
\bibliography{egbib}
}

\end{document}

%% file: tab/rank.tex

\begin{table}[h]
  \centering
  \caption{The final rank and S4 score on NVIDIA AI CITY 2021 Track4.}
  \vspace{5pt}
    \begin{tabular}{c|c|c|c}
    \hline 
Rank & Team ID & Team Name & Score \\ \hline
    1     & 76    & BaiduVIS\&SIAT & 0.9355 \\ \hline
    2     & 158   & BD    & 0.922 \\ \hline
    3     & 92    & WHU-IIP & 0.9197 \\ \hline
    4     & 90    & SIS Lab & 0.8597 \\ \hline
    5     & 153   & Titan Mizzou & 0.5686 \\ \hline
    6     & 48    & BUPT-MCPRL2 & 0.289 \\ \hline
    7     & 26    & Attention Please! & 0.2184 \\ \hline
    8     & 154   & Alchera & 0.1418 \\ \hline
    9     & 12    & CET   & 0.1401 \\ \hline
    \end{tabular}%
  \label{rank}%
\end{table}%